\documentclass[conference]{IEEEtran}
\usepackage[utf8]{inputenc}
\usepackage{cite}
\usepackage{amsmath,amssymb,amsfonts}
\usepackage{algorithmic}
\usepackage{graphicx}
\usepackage{textcomp}
\usepackage{xcolor}
\def\BibTeX{{\rm B\kern-.05em{\sc i\kern-.025em b}\kern-.08em
    T\kern-.1667em\lower.7ex\hbox{E}\kern-.125emX}}
    
\usepackage{amsmath,amssymb,amsthm}
\usepackage{mathrsfs}
\usepackage{algorithm}
\usepackage{algorithmic}
\usepackage{caption}
\usepackage{ltablex}
\usepackage{booktabs}
\usepackage{hyperref}
\usepackage{subcaption}
\usepackage{color}
\usepackage{graphicx}

\usepackage{enumitem}

\usepackage{array}
\usepackage{multicol}

\usepackage{caption}
\usepackage{booktabs}
\usepackage{float}
\usepackage{etoolbox}

\usepackage[bottom]{footmisc}
\usepackage{xcolor}
\usepackage{booktabs}
\usepackage{breqn}
\usepackage{array}
\usepackage{bm}

\usepackage{makecell}

\newcolumntype{L}[1]{>{\raggedright\let\newline\\\arraybackslash\hspace{0pt}}m{#1}}
\newcolumntype{C}[1]{>{\centering\let\newline\\\arraybackslash\hspace{0pt}}m{#1}}
\newcolumntype{R}[1]{>{\raggedleft\let\newline\\\arraybackslash\hspace{0pt}}m{#1}}
\newcommand{\bx}{\ensuremath{\mathbf{x}}}

\newcommand{\bz}{\ensuremath{\mathbf{z}}}

\newcommand{\brisk}{\ensuremath{\ddot{\mathbf{risk}}}}
\newcommand{\dbx}{\ensuremath{\ddot{\mathbf{x}}}}
\newcommand{\cbx}{\ensuremath{\check{\mathbf{x}}}}

\definecolor{dgreen}{RGB}{63, 175, 115}

\title{Personalized Cardiovascular Disease Risk Mitigation via Longitudinal Inverse Classification}
\author{\IEEEauthorblockN{Michael T.~Lash}
\IEEEauthorblockA{\textit{Business Analytics Area} \\
\textit{University of Kansas}\\
Lawrence, USA \\
michael.lash@ku.edu}
\and
\IEEEauthorblockN{W. Nick Street}
\IEEEauthorblockA{\textit{Business Analytics Department} \\
\textit{University of Iowa}\\
Iowa City, USA \\
nick-street@uiowa.edu}
}

\begin{document}

\maketitle

\begin{abstract}
    Cardiovascular disease (CVD) is a serious illness affecting millions world-wide and is the leading cause of death in the US. Recent years, however, have seen tremendous growth in the area of personalized medicine, a field of medicine that places the patient at the center of the medical decision-making and treatment process. Many CVD-focused personalized medicine innovations focus on genetic biomarkers, which provide person-specific CVD insights at the genetic level, but do not focus on the practical steps a patient could take to mitigate their risk of CVD development. In this work we propose longitudinal inverse classification, a recommendation framework that provides personalized lifestyle recommendations that minimize the predicted probability of CVD risk. Our framework takes into account historical CVD risk, as well as other patient characteristics, to provide recommendations. Our experiments show that earlier adoption of the recommendations elicited from our framework produce significant CVD risk reduction.
\end{abstract}

\section{Introduction}

Cardiovascular disease (CVD) is a serious illness that affects millions in both the United States and across the world. In 2016, CVD was the leading cause of death of in the US, responsible for more than 840,000 deaths \cite{benjamin2019heart}, thus accounting for one in three of all US deaths \cite{american2019heart}. Moreover, over 1 million individuals experienced a CVD event in 2019. Comprehensively, these statistics paint a troubling picture of the current state of cardiovascular health. 

Encouragingly, however, over the past four to five decades cardiovascular disease mortality has been on the decline \cite{mensah2017decline}. This observed decline is attributable to many advances in modern medicine, ranging from surgical bypass innovations to preventive care and education. Most recently, however, researchers have begun to worry that CVD mortality may be plateauing and the decline even reversing among certain populations \cite{mensah2017decline}.

Advances in precision medicine may be the key to encouraging continued CVD mortality reduction.
Precision medicine methodologies operate by placing the individual at the center of the medical decision-making process to find the interventions or treatments that maximize the probability of a good outcome. Therefore, these methodologies consider person-specific characteristics that have a bearing on the medical issue at hand. Such characteristics may be incredibly fine-grained, defined in terms of one's genetic make-up, or more course-grained, such as demographic information, and may also include factors that pertain to one's lifestyle, such as exercise and eating habits. 

A particularly promising stream of precision medicine research is in the area of machine learning. Broadly speaking, machine learning methods induce (i.e., train) a model that learns a mapping from features (i.e.,~variables) to some outcome of interest, either continuously valued (regression) or discrete (classification) using historical data. In the context of CVD precision medicine, the features represent patient characteristics (demographics, lifestyle), which are mapped to CVD outcome, and the historical data are historical medical records. The trained model can then be used to make predictions as to the disease outcome for new patients.

Trained machine learning models are incapable of providing personalized (i.e.,~precision) recommendations that mitigate the probability of developing the disease they've learned to predict, however. In other words, machine learning models are strictly diagnostic. To turn predictions into precision prescriptions additional innovations are needed.

Inverse classification turns predictions into personalized prescriptions by finding the instance-specific (patient-specific) feature value perturbations that minimize the probability of an undesirable classification. These perturbations correspond to manipulations of the patient's characteristics. Intuitively, not all characteristics can be changed. For instance, it would make little sense to perturb age or genetic characteristics because the patient cannot change her age or genetics. On the other hand, she can change her eating habits and exercise levels. Therefore, perturbations are only made to lifestyle or similarly manipulable feature values. The perturbations are moderated by feasibility constraints that can be tailored to the individual and are implemented to avoid infeasible recommendations (e.g., run 1000 miles/day). Ultimately, the perturbations obtained from application of inverse classification represent personalized recommendations that optimally minimized the probability (risk) of CVD.

In this work we propose a \textit{longitudinal} inverse classification approach to CVD risk minimization. Longitudinal, in this context, indicates that we will observe the same patients over several time periods. By devising a longitudinal approach to the inverse classification problem we can observe how inverse classification-elicited recommendations affect CVD risk over multiple, temporal periods. In other words, we can explore how personalized recommendations adopted earlier vs. later affect future CVD risk over several periods of time. The recommendations elicited from the proposed inverse classification process will be in the form of lifestyle adjustments, such as changes to diet and exercise. The proposed longitudinal approach will also allow us to more realistically model CVD development by taking into account past patient history.



Our innovations in the areas of CVD risk minimization, precision medicine, and inverse classification can be summarized as follows: \textbf{(1)} We develop an updated inverse classification framework to accommodate longitudinal data, thereby creating a more realistic formulation that creates temporal linkages between each observation period. \textbf{(2)} Using our proposed longitudinal inverse classification framework we show how early adoption of the elicited precision recommendations continues to lower CVD development probability during subsequent periods. In other words, we show how lifestyle adjustments made earlier continue to reduce CVD probability during later observation periods. \textbf{(3)} We explore and solve an intermediary ``missing variable'' problem that frequently arises when longitudinal data are used. Namely, we propose and evaluate the use of so-called ``missing value estimators'' that provide estimates for features that are not measured during some time periods in the longitudinal study.

The paper proceeds by first disclosing related work in Section 2, followed by our \textit{longitudinal inverse classification} formulation in Section 3. Section 4 discloses our data, experiments and results and Section 5 concludes the paper.

\section{Related Work}

We decompose our discussion of related works according to four categories: precision and personalized medicine, personalized medicine applied to CVD, machine learning with longitudinal data, and inverse classification.

Personalized and precision medicine are relatively new avenues of research that have been gaining popularity in recent years. Personalized medicine differs from so-called ``traditional'' forms of one-size-fits-all medicine by placing the individual at the center of medical decision-making process to find a medical course of action that is specific to the individual \cite{schork2015personalized}. Personalized medicine may operate at a very fine-grained genetic level to discover new drugs \cite{cardon2016precision}, for instance, or more course-grained demographic and lifestyle levels to prevent and mitigate chronic diseases \cite{minich2013personalized}. This study is concerned with the latter, where we are attempting to mitigate the risk of CVD development.

Personalized medicine research in the area of CVD is somewhat understudied, particularly in regard to technologies that have been deployed and are being used in practice \cite{persmedheart2018}. Nevertheless, recent years have witnessed a growth in personalized medicine for CVD \cite{lee2012personalized}, much of which is focused on genomic research with particular emphasis being placed on the identification of biomarkers indicative of the disease \cite{leopold2018emerging, lee2012personalized, currie2018precision, kahn2008impact}. In contrast, this work focuses on the creation of a framework that can immediately be used to make personalized lifestyle recommendations that mitigate long-term CVD risk, thus bridging the gap between traditional preventive lifestyle-focused CVD research \cite{benjamin2019heart,american2019heart} and personalized medicine.

Emergent data mining and machine learning research involving \textbf{longitudinal data} is focused on methodologically leveraging such data, as well as the specific domains in which such methods can be employed. In \cite{Razavian2016} the authors explore deep neural networks that, with minimal preprocessing, learn a mapping from patients' lab tests to over 130 diseases (multi-task learning). In \cite{Wang2014} unsupervised learning methods are applied to longitudinal health data to learn a disease progression model. The model can subsequently be used to aid patients in making long-term treatment decisions. These works exemplify the way in which models can be learned to aid in predicting disease \cite{Razavian2016} and in forecasting disease progression \cite{Wang2014}. In this work we examine how (1) coupling longitudinal data with predictive models can make disease risk estimation more accurate and (2) how predictive models that incorporate historical risk and past behavior can be used to make recommendations that optimally minimize the likelihood of developing a certain disease.

\textbf{Inverse classification} methods are varied in their approach to finding optimal recommendations, either adopting a greedy \cite{Aggarwal2010,Chi2012,Mannino2000,Yang2012} or non-greedy formulation \cite{Barbella2009,Pendharkar2002,Lash2017a,Lash2017b}. Past works also vary in their implementation of constraints that lead to more realistic recommendations, either being completely unconstrained \cite{Aggarwal2010,Chi2012,Yang2012}, or constrained \cite{Barbella2009,Pendharkar2002,Mannino2000,Lash2017a,Lash2017b}. In this work, we adopt the formulation and framework related by \cite{Lash2017a,Lash2017b} which accounts for (a) the features that can and cannot be changed (e.g.~age cannot be changed, but exercise levels can), (b) varying degrees of change difficulty (feature-specific costs) and (c) a restriction on the cumulative amount of change (budget). As in \cite{Lash2017b}, we implement a method that avoids making greedy recommendations while still accounting for (a), (b) and (c).

\section{Longitudinal Inverse Classification}
In this section we begin by providing some preliminary notation followed by a definition and formulation of past CVD risk estimators and a definition and formulation of so-called missing feature value estimators, which are necessarily incorporated into our formulation of a longitudinal inverse classification framework.

\subsection{Preliminaries}

Let $\{(id_v^{(i)},\bx^{(i)}_v,y^{(i)}_{v+1})\}_{i=1}^{n_v}$ be a dataset of $n$ instances, where $id_v^{(i)}$ is a value that uniquely identifies patient $i$, $\bx^{(i)}_v$ denotes patient $i$'s feature vector, $y^{(i)}_{v+1} \in \{0,1\}$ is the known CVD outcome of patient $i$ with $1$ indicating that the patient developed CVD and $0$ indicating that the patient did not develop CVD. Furthermore, $v=1,\dots,V$ denotes the discrete temporal unit in which the $\{\bx^{(i)}_v\}_{i=1}^{n_v}$ patient characteristics were measured. On the other hand, the outcome of interest $y^{(i)}_{v+1}$ is observed at the next (i.e.,~immediately proceeding) temporal period, hence the $v+1$ notation. Note that, for the sake of notational convenience, we assume that there is a way to determine $\{y^{(i)}_{v+1}\}_{i=1}^{n_v}$ when $v=V$. Furthermore, let $f_v(\cdot)$ denote a classifier induced on training data from visit (i.e., temporal period) $v$; $f:\mathbb{R}^{n_v \times p_v}\rightarrow [0,1]$, where $n_v$ and $p_v$ denote the number of instances and number of features measured, respectively, at visit $v$. We assume that $f_v(\cdot)$ is a classification function that produces probability estimates of CVD outcome, hence the bounded output domain of $[0,1]$. This is not a limiting assumption since any classifier that does not natively produce probability estimates can be ``trained'' to produce such estimates using Platt Scaling \cite{platt1999}.

In this setting, an assumption made about the longitudinal datasets is that of \textit{instance continuity}, where the same instances are represented in each of the defined $v=1,\dots,V$ datasets. In other words, given a set of numbers uniquely identifying each instance in a dataset at visit $v$, denoted $ID_v$, the following holds
\begin{align}
    \label{eq:ids}
    ID_{v+1} = \{id_{v+1}^{(i)}\}_{i=1}^{n_{v+1}} \subseteq ID_v = \{id_v^{(i)}\}_{i=1}^{n_v}: \text{ for }v=1,\dots,V.
\end{align}
The specification given in \eqref{eq:ids} allows us to follow the progression of individual instances over time. However, to insure that the classification function induced on $\{\bx^{(i)}_{v}\}_{i=1}^{n}$ learns a consistent representation of the outcome, namely $\{y^{(i)}_v\}_{i=1}^{n}$, the instances who experience the outcome of interest at $v$ are removed from the $v+1,\dots,V$ datasets. In other words, the following holds
\begin{align}
    \label{eq:exclude}
    \{id_{v+1}^{(i)}\}_{i=1}^{n_{v+1}} \cap  \{id_v^{(i)} \vert y^{(i)}_v = 1\}_{i=1}^{n_v} = \emptyset : \text{ for }v=1,\dots,V.
\end{align}

Let $\mathcal{F}_v$ denote the feature set at visit $v$ and $U,I,D$ denote index sets corresponding to certain feature categories. The $U$ index set denotes ``unchangeable'' features, such as age and height that cannot be changed by the patient and for which we cannot provide recommendations -- e.g., $\mathcal{F}_U=\{\text{age},\text{height},\dots\}$. The $D$ index set, conversely, denotes features that can be changed ``directly'' and represent the features for which we can provide recommendations. The features include lifestyle attributes such as diet and exercise -- e.g.,~$\mathcal{F}_{D}=\{\text{diet},\text{exercise},\dots\}$. The $I$ index set indicates so-called ``indirectly'' changeable features. This set of features correspond to those variables that cannot be changed directly, but may change as a consequence to changes made in the $\mathcal{F}_{D}$ feature set. Such variables might include blood pressure and blood glucose, among others -- e.g.,~$\mathcal{F}_{I}=\{\text{blood pressure},\text{blood glucose},\dots\}$.

Since the $\mathcal{F}_I$ features change as a consequence of manipulations made to the $\mathcal{F}_{D}$, which are the features we will ultimately be making recommendations about (i.e.,~making changes to), we wish to model this dependence explicitly. However, the $\mathcal{F}_I$ features also depend upon the $\mathcal{F}_U$, which will remain static. Therefore, let $H:\mathbb{R}^{\vert U \vert + \vert D \vert} \rightarrow \mathbb{R}^{\vert I \vert}$ denote an \textit{indirectly changeable feature estimator} that takes the $\bx_U$ and $\bx_D$ feature values as input and provides estimates estimates for the $\bx_I$ feature values. Therefore, we can use $H(\cdot)$ to track how changes made to $\bx_D$ influence the $\mathcal{F}_I$ feature values. This indirect feature value estimator will be substituted in place of the $\bx_I$ feature values when we formulate our longitudinal inverse classification framework in the proceeding subsection.

\subsection{Historical Risk Estimates Features}

Since we observe the same instances in the datasets specified over the $v=1,\dots,V$ visits, we can also observe how the outcome of interest temporally develops.  Moreover, we can induce models that take into account \textit{outcome development} (i.e.,~disease progression) when learning a mapping to the outcome of interest.

To induce such models some notion of outcome development (i.e.,~progression) must exist somewhere in the data. Initially, however, the datasets are temporally ``unlinked''. In other words no connection exists to link a patient's observed features values at an earlier time to those at later time. A naive way to create a temporal link would be to carry-forward a patient's observed feature values at each of the $v^{\prime}=1,\dots,v-1$ time periods and add them as features in the $v$ period. This would, of course, create increasingly large feature vectors and is not practical.\footnote{We do, however, compare to this naive temporal linkage strategy in our experiments.} Therefore, we propose using CVD \textit{risk} estimates -- the predicted probability of CVD occurring -- as features in ``future'' datasets. The addition of such features act as temporal links between previous datasets and future datasets, providing a notion of how likely an instance was to develop CVD during previous visits, thus providing a more realistic picture of outcome development. Furthermore, CVD risk estimates are represented as single values and are therefore more practical to use than the aforementioned naive procedure.

To formalize this notion, let
\begin{align}
    risk_v^{(i)} = f_v(\bx^{(i)}_v)
\end{align}
denote an estimate of risk, obtained from classification function $f_v(\cdot)$ at visit $v$, for patient $id_v^{(i)}$ having feature values $\bx^{(i)}_v$. To create a vector of \textit{past} risk, corresponding to an arbitrary number of past visits, we can write
\begin{align}
    \label{eq:riskdef}
    \brisk^{(i)}_{v} = (f_{v-(v-1)}(\bx^{(i)}_{v-(v-1)}),f_{v-(v-2)}(\bx^{(i)}_{v-(v-2)}),\dots,\\\nonumber
    f_{v-1}(\bx^{(i)}_{v-1})).
\end{align}

\begin{figure*}
    \centering
    \includegraphics[scale=1.5]{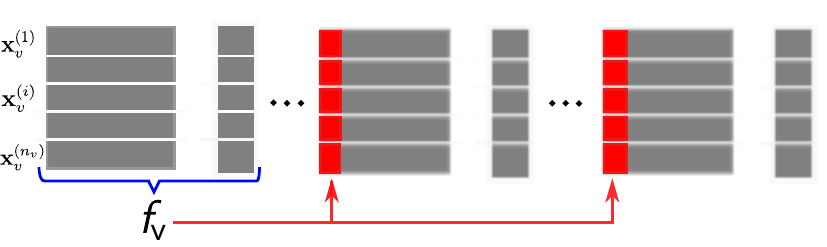}
    \caption{An illustration of the addition of past risk estimates as features in future datasets, using estimates from a single period in time only. The risk estimate features are illustrated in \textcolor{red}{red}.}
    \label{fig:risk_example}
\end{figure*}

Then, for each subsequent temporal period $v^{\prime} = v+1, \dots, V$ we construct a dataset $\{(id_{v^{\prime}}^{(i)},\dbx^{(i)}_{v^{\prime}},y^{(i)}_{v^{\prime}+1})\}_{i=1}^{n_{v^{\prime}}}$, where $\dbx^{(i)}_{v^{\prime}} = \left (\bx^{(i)}_{v^{\prime}}, \brisk^{(i)}_{v^{\prime}-1} \right )$. A simple example of this process, for a single risk estimate, is illustrated in Figure \ref{fig:risk_example}.

Correspondingly, classifiers $\ddot{f}_{v^{\prime}}(\cdot): \text{ for } v^{\prime}=v+1, \dots, V$ can be induced on the risk estimate-containing datasets.

\subsection{Missing Feature Estimators}

One issue with longitudinal data is that features measured during earlier time periods may not be measured at subsequent periods. This presents a slight problem in the context of inverse classification because we would like to induce classifiers on the same set of features from one time period to the next. More importantly, we would like to optimize over the same set of $\mathcal{F}_D$ features at each time period, observing how such optimization effects risk estimates in the future; thus we require that classifier induction take place on the same set of $\mathcal{F}_D$ features.

Therefore, we propose to use so-called \textit{missing feature-value estimators} to estimate the missing feature values at future time periods. Such estimators will be induced using the known features at visit $v$ to induce classifiers or regressors, depending upon whether the missing feature is binary or continuous. Therefore, we will define the full set of features $\mathcal{F}$ at visit $v=1$ to ensure there is at least one time period from which we can always estimate future feature values. Using these estimators, we will estimate the instances' missing feature values at visit $v$. Our proposition to use such estimators is explored in the experiments section.

To formalize this notion, let $\mathcal{F}_{v=1}$ denote the feature set for the first time period in the longitudinal study. With this in mind, the following relation holds for the data in our longitudinal context:
\begin{align}
    \label{eq:featuresub}
    \{ \mathcal{F}_{v= \tilde{v}} \subseteq \mathcal{F}_{v=1}: \tilde{v}=2,\dots,V \}.
\end{align}
The specification in \eqref{eq:featuresub} indicates that the data at visit $v=1$ is sufficiently capable of providing estimates for the missing feature values at any subsequent time period because the features used at all future time periods are subsets or proper subsets (indicating no missing features) of the features defined at $v=1$.

With this in mind, let $M_v$ denote the set of missing features at visit $v$, where $M_v = \{\mathcal{F}_{v=1} \setminus \mathcal{F}_{v}\}$. Permit the instances' subvectors at $v=1$ to be indexed using $M_v$, i.e.,~$\{\bx_{M_v,v=1}^{(i)}\}_{i=1}^{n_{v=1}}$. Then, we define the estimation of missing features, using missing feature-value estimators, as follows:
\begin{align}
    \label{eq:missfeatest}
    x_{m,v}^{(i)} = \varepsilon_{m}(\bx_{v}^{(i)}) : \text{ for } m \in {M_v}
\end{align}
where $\varepsilon_{m}(\cdot)$ denotes a classifier or regressor induced on $\{\bx_{F_v,v=1}^{(i)}\}_{i=1}^{n_{v=1}}$ to learn the mapping to each $m \in M_v$, i.e.,~$\varepsilon_{m}:\{\bx_{F_v,v=1}^{(i)}\}_{i=1}^{n_{v=1}} \rightarrow \{\bx_{m,v=1}^{(i)}\}_{i=1}^{n_{v=1}}$.

Subsequently, let
\begin{align}
    \label{eq:xmisspast}
    \cbx_v^{(i)} = \left(\dbx_{v}^{(i)},  \left(x_{m,v}^{(i)} \right)_{m \in M_v} \right)
\end{align}
where $\cbx_v^{(i)}$ indicates an instance at visit $v$ composed of the original feature values, missing feature values, and historical risk estimates. Note that we can now refer to $\cbx_v^{(i)}$ using the index sets $U,I,D$ defined in a previous subsection.

Similarly, we can induce classifiers $\check{f}_{\tilde{v}}(\cdot): \text{ for } \tilde{v}=2, \dots, V$ on data containing both past risk estimates and missing feature-value estimates.

\subsection{Inverse Classification Formulation}

With the discussion of past risk estimate and missing feature-values resolved, we can formulate our longitudinal inverse classification framework as:
\begin{align}
\label{eq:longicform}
\min_{\cbx_{D,v}} \hspace{1mm}& \check{f}_v(\cbx_{U,v},H(\cbx_{U,v},\cbx_{D,v}),\cbx_{D,v}) \\\nonumber
\text{s.t.} \hspace{1mm}& \mathbb{C}_v (\cbx_{D,v} - \bar{\cbx}_{D,v}) \leq B_v\\\nonumber
\hspace{1mm} & l_{j,v} \leq \check{x}_{j,v} \leq u_{j,v}, \text{ for } j \in D,
\end{align}
where $\mathbb{C}(\bz)=\sum_{j\in D}c_j^+(z_j)_++c_j^-(z_j)_-$, such that $(z)_+=\max\{0,z\}$ and $(z)_-=\max\{0,-z\}$, is a cost function that measures the deviation of the directly changeable feature values $\cbx_{D,v}$ from their original values $\bar{\cbx}_{D,v}$ and applies a user-specified cost $c_j:j \in D$ to these deviations. The costs $c_j$ allow for a patient to specify which feature values they would prefer to change (e.g.,~willing to exercise more than a change to diet). These cost-changes are summed together and subjected to a budget constraint $B_v$ that controls the extent of changes recommended, thus  avoiding over-radical recommendations (e.g.,~exercise 1000 hours in a week) and further allows patients and clinicians to express preferences. The $l_{j,v} (u_{j,v}):j \in D$ are lower (upper) bounds that ensure a recommendation makes sense in the real world -- e.g., prevents negative dietary intake from being recommended. Note that in \eqref{eq:longicform} $H(\cbx_{U,v},\cbx_{D,v})$ is used instead of $\cbx_{I,v}$ to account for how changes made to the directly changeable features affect the indirectly changeable features.

\section{Experiments and Results}

In this section we propose and execute several experiments that assess the viability of our proposed use of past risk estimates, missing feature value estimators, and our specified longitudinal inverse classification framework disclosed in \eqref{eq:longicform} in being able to reduce CVD risk.

To such an end, we first disclose our proposed experiments and subsequently describe our CVD data. Next, we provide some details on the specific design choices made for our classifier, optimizer, and indirect feature value estimator. Following these disclosures, we report and discuss the results of our experiments.

\subsection{Evaluation Specifics}


\subsubsection{Experiments}
We propose the following three experiments to assess the viability of our proposed methods and corresponding inverse classification framework: \textbf{(1)} To assess the viability of our proposed use of missing feature value estimators we compare the performance of various classifiers and regressors to that of a simple carry-forward procedure (baseline). The carry-forward procedure simply entails imputing the missing feature values using the instances' known feature values at $v=1$. In other words, the values are carried forward from $v=1$ to $v$. \textbf{(2)} We explore our proposed use of previous risk estimates as features in subsequent visits by inducing classifiers on data containing such estimates, inducing classifiers on data containing carried-forward past-visit instance feature vectors, and comparing the predictive performance between the two methods. The carry-forward procedure replaces our original definition of risk in \eqref{eq:riskdef} by $(\bx^{(i)}_{v-(v-1)},\dots,\bx^{(i)}_{v-1},\bx^{(i)}_v)$. \textbf{(3)} We examine the effects of applying our longitudinal inverse classification procedure to different temporal observation periods on future CVD risk. Particularly, we examine the effect of a single application of inverse classification applied early vs repeated applications of inverse classification on predicted CVD risk observed in future periods.

\subsubsection{Dataset Description}

To perform the experiments outlined in the previous subsection we use ARIC (atherosclerosis risk in communities study) data, a freely available dataset (subject to approval) from BioLINCC. Derivation of longitudinal data from this study produced $V=3$ visit datasets having the statistics outlined in Table \ref{tab:dset_sum}, with visit four being used to define the CVD outcome ($y$) associated with the third visit ($v=3$).
\begin{table}[h]
	\centering
	\begin{tabular}{|l|c|c|c|}
		\hline
		Dataset & Instances & Feats/Missing & $y_{v+1}=1$ \\
		\hline
		$v=1$ & 12223 & 122/0 & 232 \\
		\hline
		$v=2$ & 11057 & 98/24 & 249 \\
		\hline
		$v=3$ & 9883 & 74/48 & 231 \\
		\hline	
	\end{tabular}
	\caption{ Dataset descriptors.\label{tab:dset_sum}}
\end{table}
The features defined in $v=1$, along with additional dataset details, such as the $\mathcal{F}_U, \mathcal{F}_I, \mathcal{F}_D$ features can be found in the Appendix section at the end of this work.

\subsubsection{Classifier, Optimizer, and Indirect Feature Value Estimator}

Our longitudinal inverse classification framework disclosed in \eqref{eq:longicform} can be optimized by applying various gradient descent methods if $\check{f}_v(\cdot)$ is differentiable and has an $L$-Lipschitz continuous gradient. Otherwise, various heuristics such as genetic algorithms, simulated annealing, etc. can be applied. In our inverse classification experiments we use support vector machines (SVM) for $\check{f}_v(\cdot)$, which is both differentiable and has an $L$-Lipschitz continuous gradient. Thus, we use projected gradient descent (PGD) \cite{Nesterov07composite,Ghadimi:13a} to optimize the objective function in \eqref{eq:longicform} to obtain recommendations. We elect to use kernel regression as our indirect feature estimator since it is also differentiable and provided good predictive performance experimentally.

\subsection{Experiment 1: Missing Feature Estimators}

To justify our use of missing feature-value estimators, we select three known features at $v=2$ at random and treat them as if they are missing. Subsequently, we induce missing feature-value estimators with $v=1$ data, using the remaining known features at $v=2$ (i.e., sans the three randomly selected features). We then estimate the three features at $v=2$ using the induced estimators and derive predictive performance measures by comparing the estimates to the actual values (AUC for categorical features and MSE for continuous). As stated during our description of this experiment, we compare the estimators to a ``carry-forward'' procedure. The results of this experiment are disclosed in Table \ref{tab:miss_feat}.

\begin{table}[h]
	\centering
	\begin{tabular}{|l|c|c|c|}
		\hline
		 \diaghead{\theadfont Diag CHeadssss  }{Method}{Feat/type}& Alcohol/cont & Statin Use/bin & Hematocrit/cont \\
		\hline
		Carry & 50.55 & .579  & 7.55  \\
		\hline
		RBF SVM & 47.65 & .50  & 9.31  \\
		\hline
		Lin SVM & 47.65 & .50 & 9.31 \\
		\hline
		CART & 30.97 & .569  & 2.37 \\
		 \hline	
		kNN & \textbf{29.37} & .50  & 12.81 \\
		  \hline
		Log Reg & NA & \textbf{.984}  & NA \\
		   \hline
		Ridge & \textbf{29.42}* & NA & \textbf{1.24} \\
		    \hline
	\end{tabular}
	\caption{ Missing feature estimation results given in MSE for continuous features and AUC for binary.\label{tab:miss_feat}}
\end{table}

Table \ref{tab:miss_feat} first shows that, for each of the three features, there was at least one estimator that outperformed  the ``carry-forward'' procedure, thus justifying the use of estimators over the naive carry-forward procedure. Second, we observe that ridge regression is the best performing continuous value estimator for ``hematocrit'' and a very close second for ``alcohol consumed''. Therefore, in our inverse classification experiments, we used ridge regression to estimate continuous features. Finally, we found that logistic regression was best able to estimate the categorical feature ``statin use'' and thus use logistic regression to estimate categorical features in our inverse classification experiments.

\subsection{Experiment 2: Risk Estimates vs.~Feature Vectors}

\begin{figure}[!htp]
    \centering
    \includegraphics[scale=0.80]{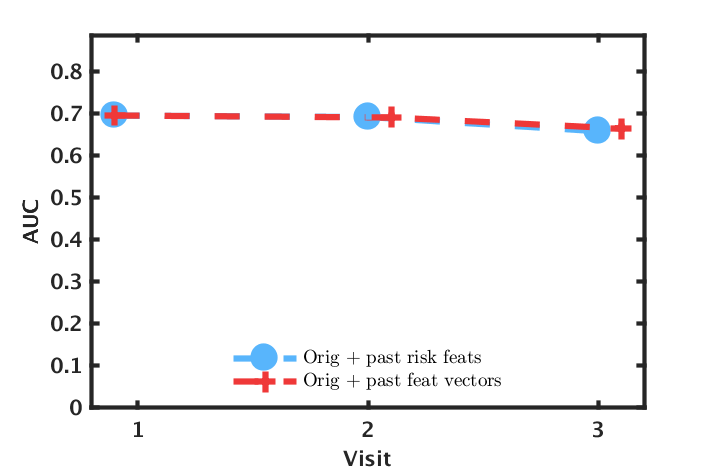}
    \caption{Performance of classifiers induced using past risk estimates vs. classifiers induced using historical feature vectors, by visit. }
    \label{fig:comparisonres}
\end{figure}

To create a temporal linkage between longitudinal datasets, we proposed to use past risk estimates as features in future visit datasets. To assess the viability of this method we induce classifiers on datasets containing these features, also inducing classifiers on datasets containing feature vectors that are carried forward from past visits. We then compare the predictive performance between the two methods to assess the viability of our proposed strategy. The results of this comparison-based experiment are presented in Figure \ref{fig:comparisonres}.

Figure \ref{fig:comparisonres} shows that the predictive performance of the two methods is incredibly similar. At $v=3$, the carry-forward procedure enjoys a $<.01$ AUC benefit over our proposed past risk method. Since the two methods perform virtually identical we elect to adopt our proposed past risk method, which provides a much more compact representation than the carry-forward procedure (i.e., past risk represents each past dataset as a single value, whereas the carry-forward procedure uses the entire feature vector of each past visit).

\subsection{Experiment 3: Longitudinal Inverse Classification}

To examine the effects of longitudinal inverse classification, we execute the inverse classification procedure in two different ways:
\begin{itemize}
\item[a.] We execute the inverse classification procedure at $v=1$. We then use these optimized $D$ feature-values as the $D$ feature values at $v=2$ and $v=3$ and predict their probability of CVD at each of these visits. This represents instances' continued implementation of their personalized recommendation from $v=1$. Additionally, we use estimates of historical risk at $v=2$ and $v=3$ based on the optimized $D$ values at $v=1$. We report the results in terms of average probability of CVD.
\item[b.] We execute the same procedure as in (a.), above, but also apply a similar procedure at $v=2$ using the newly optimized $D$ feature-values as the $D$ feature-values for the corresponding instances in the $v=3$ dataset. This represents instances' that make initial improvements toward a beneficial outcome, but then also make additional, follow-up improvements in an attempt to further reduce their probability of CVD. We also report these results in terms of average probability of CVD.
\end{itemize}

\begin{figure}[!htp]
    \centering
    \includegraphics[scale=0.59]{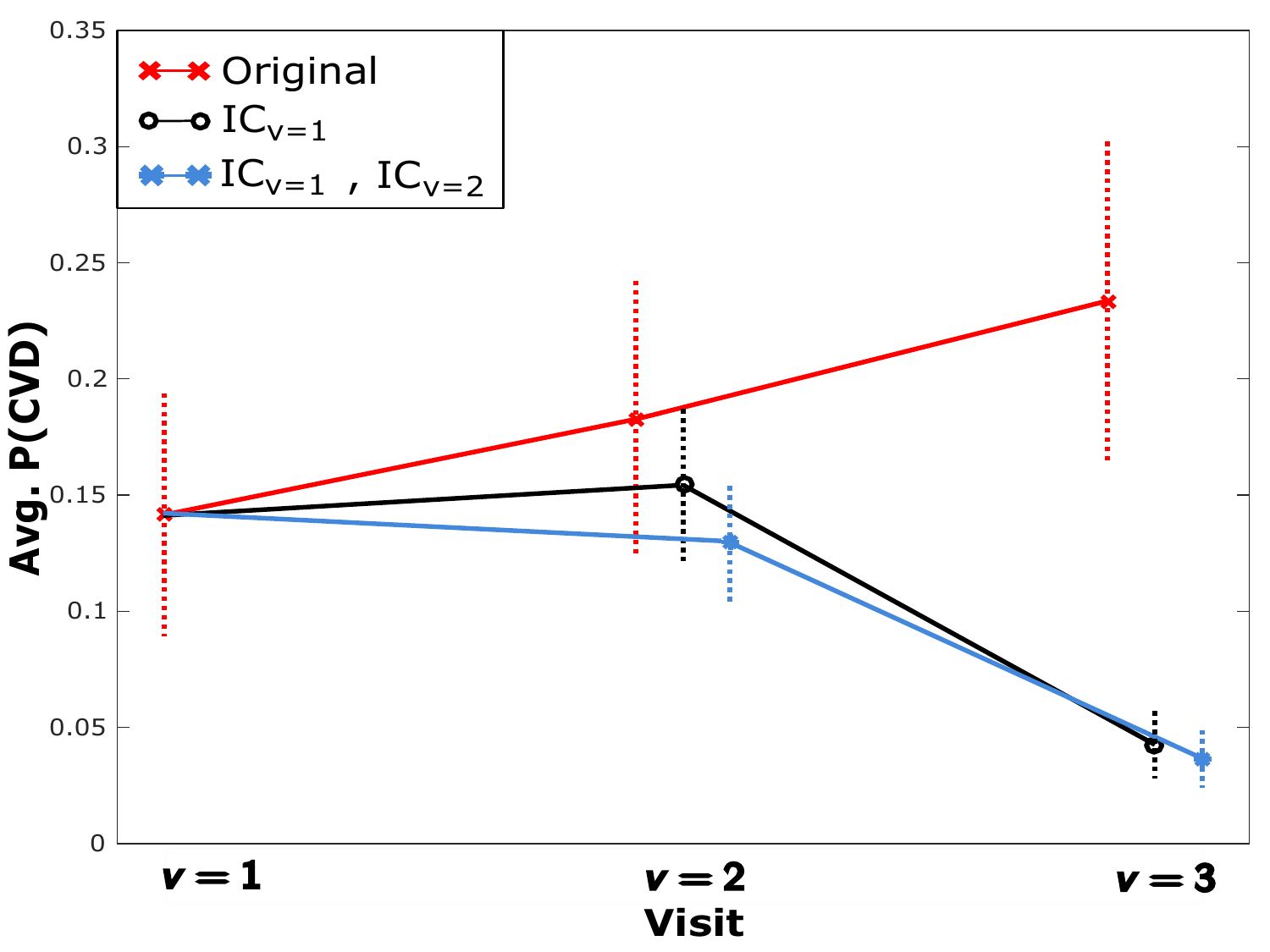}
    \caption{Longitudinal inverse classification results by visit with a budget $B=2$. \textcolor{red}{Red} represents average probability of CVD without applying inv. Black indicates the average probability results after applying inverse classification at $v=1$. \textcolor{blue}{blue} indicates the average probability results after applying inverse classification at $v=1$ and $v=2$.}
    \label{fig:longicres}
\end{figure}

Figure \ref{fig:longicres} shows the results of (a.) in black and (b.) in \textcolor{blue}{blue}, which can be compared to the original predicted probability of CVD in \textcolor{red}{red}.

Examining the results of both (a.) and (b.) at $v=2$, we can see that both were able to reduce the average probability of CVD; the result of (b.), expressed in \textcolor{blue}{blue}, suggests that a second application of the inverse classification process further reduces the average probability of CVD.

Interestingly, however, the results of (a.) and (b.) at $v=3$ show that, while (a.) and (b.) kept the average predicted probability of CVD low, their difference in average predicted outcome is very comparable ((b.) is slightly less than (a.)). This is surprising because it suggests that applying a sufficient amount of initial effort to improve ones probability of a good outcome has a comparable ``long-term'' effect to application of additional CVD-mitigating effort later on, perhaps indicating/reinforcing the notion that interventions are best made earlier rather than later (by virtue of the ``diminishing effect'' observed).

\section{Conclusions}

Cardiovascular disease is a major health concern world wide. Personalized medicine provides an avenue by which CVD risk can be mitigated, thus potentially improving the health outlook for millions. Much of the on-going CVD research on personalized medicine focuses on the discovery of genetic biomarkers, but does not address personalized lifestyle adjustments that may reduce CVD risk. In this work, we proposed longitudinal inverse classification that makes personalized lifestyle recommendations that mitigate the risk of CVD, also addressing the sub-problems of missing features and the creation of temporal linkages among longitudinal datasets. We show that our proposed methods to address these two subproblems are reasonable and that longitudinal inverse classification can be used to reduce the long-term risk of CVD. Furthermore, we found that personalized lifestyle adjustments adopted earlier are approximately as beneficial as repeated lifestyle adjustment made over several time periods.

\bibliographystyle{IEEEtran}
\bibliography{long}

\begin{thebibliography}{10}
\providecommand{\url}[1]{#1}
\csname url@samestyle\endcsname
\providecommand{\newblock}{\relax}
\providecommand{\bibinfo}[2]{#2}
\providecommand{\BIBentrySTDinterwordspacing}{\spaceskip=0pt\relax}
\providecommand{\BIBentryALTinterwordstretchfactor}{4}
\providecommand{\BIBentryALTinterwordspacing}{\spaceskip=\fontdimen2\font plus
\BIBentryALTinterwordstretchfactor\fontdimen3\font minus
  \fontdimen4\font\relax}
\providecommand{\BIBforeignlanguage}[2]{{%
\expandafter\ifx\csname l@#1\endcsname\relax
\typeout{** WARNING: IEEEtran.bst: No hyphenation pattern has been}%
\typeout{** loaded for the language `#1'. Using the pattern for}%
\typeout{** the default language instead.}%
\else
\language=\csname l@#1\endcsname
\fi
#2}}
\providecommand{\BIBdecl}{\relax}
\BIBdecl

\bibitem{benjamin2019heart}
E.~J. Benjamin, P.~Muntner, and M.~S. Bittencourt, ``Heart disease and stroke
  statistics-2019 update: a report from the american heart association,''
  \emph{Circulation}, vol. 139, no.~10, pp. e56--e528, 2019.

\bibitem{american2019heart}
A.~H. Association \emph{et~al.}, ``Heart disease and stroke statistics 2019
  at-a-glance,'' \emph{online}, 2019.

\bibitem{mensah2017decline}
G.~A. Mensah, G.~S. Wei, P.~D. Sorlie, L.~J. Fine, Y.~Rosenberg, P.~G.
  Kaufmann, M.~E. Mussolino, L.~L. Hsu, E.~Addou, M.~M. Engelgau \emph{et~al.},
  ``Decline in cardiovascular mortality: possible causes and implications,''
  \emph{Circulation research}, vol. 120, no.~2, pp. 366--380, 2017.

\bibitem{schork2015personalized}
N.~J. Schork, ``Personalized medicine: time for one-person trials,''
  \emph{Nature}, vol. 520, no. 7549, pp. 609--611, 2015.

\bibitem{cardon2016precision}
L.~R. Cardon and T.~Harris, ``Precision medicine, genomics and drug
  discovery,'' \emph{Human molecular genetics}, vol.~25, no.~R2, pp.
  R166--R172, 2016.

\bibitem{minich2013personalized}
D.~M. Minich and J.~S. Bland, ``Personalized lifestyle medicine: relevance for
  nutrition and lifestyle recommendations,'' \emph{The Scientific World
  Journal}, vol. 2013, 2013.

\bibitem{persmedheart2018}
``Taking personalized medicine to heart,'' \emph{Nature Medicine}, vol.~24,
  no.~2, p. 113, 2018.

\bibitem{lee2012personalized}
M.-S. Lee, A.~J. Flammer, L.~O. Lerman, and A.~Lerman, ``Personalized medicine
  in cardiovascular diseases,'' \emph{Korean circulation journal}, vol.~42,
  no.~9, pp. 583--591, 2012.

\bibitem{leopold2018emerging}
J.~A. Leopold and J.~Loscalzo, ``Emerging role of precision medicine in
  cardiovascular disease,'' \emph{Circulation research}, vol. 122, no.~9, pp.
  1302--1315, 2018.

\bibitem{currie2018precision}
G.~Currie and C.~Delles, ``Precision medicine and personalized medicine in
  cardiovascular disease,'' in \emph{Sex-Specific Analysis of Cardiovascular
  Function}.\hskip 1em plus 0.5em minus 0.4em\relax Springer, 2018, pp.
  589--605.

\bibitem{kahn2008impact}
R.~Kahn, R.~M. Robertson, R.~Smith, and D.~Eddy, ``The impact of prevention on
  reducing the burden of cardiovascular disease,'' \emph{Circulation}, vol.
  118, no.~5, pp. 576--585, 2008.

\bibitem{Razavian2016}
N.~Razavian, J.~Marcus, and D.~Sontag, ``Multi-task prediction of disease
  onsets from longitudinal lab tests,'' \emph{arXiv preprint arXiv:1608.00647},
  2016.

\bibitem{Wang2014}
X.~Wang, D.~Sontag, and F.~Wang, ``Unsupervised learning of disease progression
  models,'' in \emph{Proceedings of the 20th ACM SIGKDD international
  conference on Knowledge discovery and data mining}.\hskip 1em plus 0.5em
  minus 0.4em\relax ACM, 2014, pp. 85--94.

\bibitem{Aggarwal2010}
C.~C. Aggarwal, C.~Chen, and J.~Han, ``{The inverse classification problem},''
  \emph{Journal of Computer Science and Technology}, vol.~25, no. May, pp.
  458--468, 2010.

\bibitem{Chi2012}
\BIBentryALTinterwordspacing
C.~L. Chi, W.~N. Street, J.~G. Robinson, and M.~A. Crawford, ``{Individualized
  patient-centered lifestyle recommendations: An expert system for
  communicating patient specific cardiovascular risk information and
  prioritizing lifestyle options},'' \emph{Journal of Biomedical Informatics},
  vol.~45, no.~6, pp. 1164--1174, 2012. [Online]. Available:
  \url{http://dx.doi.org/10.1016/j.jbi.2012.07.011}
\BIBentrySTDinterwordspacing

\bibitem{Mannino2000}
M.~V. Mannino and M.~V. Koushik, ``{The cost minimizing inverse classification
  problem : A algorithm approach},'' \emph{Decision Support Systems}, vol.~29,
  no.~3, pp. 283--300, 2000.

\bibitem{Yang2012}
C.~Yang, W.~N. Street, and J.~G. Robinson, ``{10-year CVD risk prediction and
  minimization via inverse classification},'' in \emph{Proceedings of the 2nd
  ACM SIGHIT symposium on International health informatics - IHI '12}, 2012,
  pp. 603--610.

\bibitem{Barbella2009}
D.~Barbella, S.~Benzaid, J.~Christensen, B.~Jackson, X.~V. Qin, and
  D.~Musicant, ``{Understanding support vector machine classifications via a
  recommender system-like approach},'' in \emph{Proceedings of the
  International Conference on Data Mining}, 2009, pp. 305--11.

\bibitem{Pendharkar2002}
P.~C. Pendharkar, ``A potential use of data envelopment analysis for the
  inverse classification problem,'' \emph{Omega}, vol.~30, no.~3, pp. 243--248,
  2002.

\bibitem{Lash2017a}
M.~T. Lash, Q.~Lin, W.~N. Street, J.~G. Robinson, and J.~Ohlmann, ``Generalized
  inverse classification,'' in \emph{2017 SIAM International Conference on Data
  Mining (SDM)}.\hskip 1em plus 0.5em minus 0.4em\relax SIAM, 2017, pp.
  162--170.

\bibitem{Lash2017b}
M.~T. Lash, Q.~Lin, W.~N. Street, and J.~G. Robinson, ``A budget-constrained
  inverse classification framework for smooth classifiers,'' in \emph{Data
  Mining Workshops (ICDMW), 2017 IEEE International Conference on}.\hskip 1em
  plus 0.5em minus 0.4em\relax IEEE, 2017, pp. 1184--1193.

\bibitem{platt1999}
J.~Platt, ``{Probabilistic outputs for support vector machines and comparisons
  to regularized likelihood methods},'' \emph{Advances in Large Margin
  Classifiers}, vol.~10, no.~3, pp. 61--74, 1999.

\bibitem{Nesterov07composite}
Y.~Nesterov, ``Gradient methods for minimizing composite objective function,''
  \emph{Mathematical Programming, Series B}, vol. 140, pp. 125--161, 2013.

\bibitem{Ghadimi:13a}
S.~Ghadimi and G.~Lan, ``Stochastic first- and zeroth-order methods for
  nonconvex stochastic programming,'' \emph{SIAM Journal on Optimization},
  vol.~23, pp. 2341--2368, 2013.

\end{thebibliography}
\renewcommand{\tablename}{ST.}
\setcounter{table}{0}

\section*{Appendix}
The following tables enumerate the variables used in this study at $v=1$.

\begin{table}[!htbp]
	\centering
	\begin{tabular}{|C{6.4cm}|}
		\hline
		\textbf{Feature Name} \tabularnewline
		\hline
		Insulin (uu-ml), Height (cm), Age, Peripheral Artery Disease, Peripheral Artery Disease (definition 2), 
		Plaque/shadowing in either internal, Plaque in either internal carotid, Cholesterol lowering med (last 2 weeks),
		Hypertension (definition 5), Education level, Diabetes, Age when menopause began, Menopause status, Ever smoked cigarettes,
		High blood pressure med (past 2 weeks), Agina-chest pain med (past 2 weeks), Heart rhythm control med (past 2 weeks),
		Heart failure med (past 2 weeks), Blood thinning med (past 2 weeks), Blood sugar med (past 2 weeks), Stroke med (past 2 weeks),
		Walking leg pain med (past 2 weeks), Headache or cold med (past 2 weeks), Pain meds (past 2 weeks), Gender, Race, Years smoked cigarettes \tabularnewline
		\hline
	\end{tabular}
	\caption{Unchangeable features $\mathcal{F}_U$ for the ARIC CVD dataset.}
	\label{tab:cvd_unchange}
\end{table}

\begin{table}[!htbp]
	\centering
\centering
	\begin{tabular}{|C{6.4cm}|} 
		\hline
		\textbf{Feature Name: $\sigma$} \tabularnewline
		\hline
		BMI (Body Mass Index), Recalibrated HDL cholesterol (mg/dl), Re-calibrated LDL cholesterol (mg/dl),Total cholesterol (mmol/L), Total triglycerides (mmol/L), 2nd and 3rd systolic blood pressure (avg.), 2nd and 3rd systolic blood pressure (avg.) Num 2, Waist girth (cm), Hip girth (cm), Heart rate, White blood count,
		Apolipoprotein AI(mg-dl), Apolipoprotein B (mg-dl), Apolp(A) Data (ug-ml), Ankle-brachial index (Def 4), FV(1)/FVC Predicted (\%), FEV(1) (L), FVC (L), 
		Hematocrit, Hemaglobin, Platelet count, Neutrophils, Neutrophil bands, Lymphocytes, Monocytes, Eosinophils, Basophils, APTT Value, VIII: C Value, Fibrinogen Value, VII Value, ATIII Value, Protein: C Value, VWF Value, Cornell voltage (uV), Waist-hip ratio, Vegetable fat (\% kcal), Carbs (\% kcal), Alcohol (\% kcal), Omega fatty acid (g), Calf girth (cm), Subcaps measure 2 (mm), Triceps measure 2 (mm), 
		Uric acid (mg-dl), Total protein (gm-dl), Albium (gm-dl), Phosphorus (mg-dl), Magnesium (meq-l), Calcium (mg-dl), Urea nitgrogen (mg-dl), Potassium (mmol-l), 
		Sodium (mmol-l), Creatinine (mg-dl), Weight (lb),  Total fat (\% kcal), Saturate fatty acid (\% kcal), Protein (\% kcal), Polyunsaturated fatty acid (\% kcal), Monounsaturated fatty acid (\% kcal), Total fat (g) \tabularnewline
		\hline
	\end{tabular}

	\caption{Indirectly changeable features $\mathcal{F}_I$ for the ARIC CVD dataset.}
	\label{tab:bench_indirchange}
\end{table}

\begin{table}[H]
	\centering
	\begin{tabular}{|C{.8cm}|C{6.4cm}|}
		\hline
		$\mathbf{c^{+}/c^{-}}$ & \textbf{Feature:Cost} \tabularnewline
		\hline
		$c^{+}$ & Dark or grain breads: 3, Peanut butter: 4, Nuts: 5, Other(prunes,avocado): 5, Vegetables: 6, Fruit: 6, Fiber: 7, Vegetable fat: 5, Polyunsaturated fat: 5 \tabularnewline
		\hline
		$c^{-}$ & Liver: 8, White carbs: 6, Fish: 9, Cereal: 4, Cigarettes: 9, Caffeine: 7, Carbs: 7, Cholesterol: 6, Sodium: 7, Animal fat: 7, Saturated fat: 6 \tabularnewline
		\hline
		$c^{+}/c^{-}$ & Exercise hours: 10, Alcohol: 9 \tabularnewline
		\hline
	\end{tabular}\caption{Directly changeable features $\mathcal{F}_D$ for the ARIC CVD dataset.}
	\label{tab:aric_changeable}
\end{table}

\end{document}